\documentclass[runningheads]{llncs}
\usepackage[T1]{fontenc}
\usepackage{amssymb}
\usepackage{marvosym}
\usepackage{orcidlink}

\usepackage{graphicx,verbatim}
\usepackage{graphicx}
\usepackage{float}
\usepackage{booktabs}
\usepackage{bm}
\usepackage{hyperref}
\begin{document}
\title{ProBAG: Prototype-Guided Boundary-Aware Graph Diffusion for Weakly Supervised Histopathology Segmentation}
\titlerunning{ProBAG for Weakly Supervised Histopathology Segmentation}

%

\author{
Duy-Dong Nguyen\inst{1} \and
Le-Van Thai\inst{1} \and
Hoai Nhan Pham\inst{1} \and
Ngoc Lam Quang Bui\inst{2} \and
Tam Tran\inst{6}\textsuperscript{(\Letter)}\orcidlink{0009-0000-9702-425X}  \and
Zhi Huang\inst{7}\textsuperscript{(\Letter)}\orcidlink{0000-0001-6982-8285}
}

\authorrunning{Duy-Dong Nguyen et al.}

\institute{
AI VIETNAM Lab, Vietnam\\
\and
Department of Mechanical System Engineering, Jeonbuk National University, Republic of Korea\\
\and
John T. Milliken Department of Medicine, Washington University School of Medicine, Saint Louis, MO, USA\\
\email{tamt@wustl.edu}
\and
Perelman School of Medicine, University of Pennsylvania, USA\\
\email{zhi.huang@pennmedicine.upenn.edu}
}

\maketitle

\begin{abstract}
Weakly supervised semantic segmentation enables histopathology tissue segmentation from image-level annotations, avoiding costly pixel-level labeling by expert pathologists. However, CAM-based methods often localize only highly discriminative regions and remain unreliable near tissue interfaces. We propose ProBAG, a stage-1 pseudo-mask generator that combines dataset-specific visual prototypes with pathology-aligned CONCH text prototypes over multi-scale frozen UNI features. ProBAG introduces two complementary mechanisms: class-wise power recalibration that reshapes inter-class competition while preserving the total foreground activation mass at each pixel, and one-step graph diffusion in which feature affinities are penalized by a late-transformer attention-context discrepancy used as a soft structural boundary cue. The resulting stage-1 pseudo-masks require neither CRF nor an external segmentation model; for complete two-stage comparison, they additionally supervise a downstream Phikon-FPN segmenter. Experiments on BCSS-WSSS and LUAD-HistoSeg show consistent gains over recent WSSS approaches, while ablations indicate that pathology-aligned text semantics provide the largest improvement and graph refinement provides a smaller complementary gain. The code is available at: 
\url{https://github.com/wterrr/WSSS}


\end{abstract}
\section{Introduction}
Histopathology image segmentation is a core task in computational pathology, supporting quantitative analysis of tumor, stroma, lymphocyte, and necrosis regions in hematoxylin and eosin (H\&E) images. Although deep learning has advanced digital pathology, dense semantic segmentation still requires costly pixel-level annotation by expert pathologists \cite{10.1093/bioinformatics/btz083,Campanella2019ClinicalgradeCP}. Weakly supervised semantic segmentation (WSSS) reduces this burden by learning pseudo-masks from cheaper annotations such as image-level tissue labels\cite{minaee2020imagesegmentationusingdeep}.

Most image-level WSSS methods rely on class activation maps (CAMs), which localize discriminative regions for classifier predictions \cite{Selvaraju_2019,zhou2015learningdeepfeaturesdiscriminative}. Prior work has improved CAM quality through seed expansion, affinity propagation, inter-pixel relations, equivariance constraints, transformer attention, and class re-activation \cite{ahn2019weaklysupervisedlearninginstance,ahn2018learningpixellevelsemanticaffinity,chen2022classreactivationmapsweaklysupervised,kolesnikov2016seedexpandconstrainprinciples,ru2022learningaffinityattentionendtoend,wang2020selfsupervisedequivariantattentionmechanism}. However, CAM supervision remains indirect, as classifiers often attend only to the most discriminative regions rather than capturing full object extent \cite{lee2022weaklysupervisedsemanticsegmentation,xie2022crosslanguageimagematching}. This limitation is particularly severe in histopathology, where tissues exhibit strong visual similarity and substantial intra-class morphological variation \cite{tang2025prototypebasedimagepromptingweakly}. As a result, CAM-derived pseudo-masks are often incomplete, class-biased, and unreliable near boundaries.

Recent histopathology WSSS studies incorporate stronger domain priors to address these challenges. WSSS4LUAD establishes lung adenocarcinoma tissue segmentation from patch-level labels \cite{han2022wsss4luadgrandchallengeweaklysupervised}, while PistoSeg exploits dataset synthesis and feature consistency \cite{Fang2023WeaklySupervisedSS}. TPRO introduces text prompting for tissue segmentation \cite{Zhang2023TPROTW}, and prototype-based prompting leverages visual prototype banks to mitigate inter-class homogeneity and intra-class heterogeneity \cite{tang2025prototypebasedimagepromptingweakly}. These methods show that semantic and prototype priors significantly improve CAM localization. However, two challenges remain: class-dependent activation imbalance in fused CAMs, where highly activated regions dominate foreground prediction, and the lack of boundary-aware refinement beyond feature-similarity propagation in prototype-based methods.

Existing prototype-based WSSS methods primarily improve semantic localization, but two failure modes remain insufficiently separated. First, independently normalized class CAMs can exhibit markedly different activation distributions, allowing broad or sharply activated tissues to dominate pixel-wise competition. Second, feature-only affinity propagation can leak across interfaces between adjacent tissues with similar local appearance. ProBAG targets these failures separately through foreground-mass-preserving class recalibration and attention-context-regularized graph propagation.

Recent advances in pathology foundation models and vision-language models provide useful cues for addressing these challenges. Frozen vision transformers can provide dense multi-scale representations, while pathology-aligned vision-language models offer semantically meaningful tissue embeddings \cite{chen2024uni,dosovitskiy2021imageworth16x16words,lu2024avisionlanguage}. These developments motivate a WSSS framework that jointly exploits visual prototypes, pathology text priors, and foundation-model attention cues. Recent studies have also demonstrated the utility of pathology foundation models for histopathology segmentation \cite{fu2025multimodalprototypealignmentsemisupervised,Van_Thai_2026_CVPR}.

Accordingly, our contribution lies not in introducing prototypes or graph propagation in isolation, but in coordinating pathology-aligned semantic prototypes with two explicit corrections for class competition and cross-interface diffusion. Our main contributions are as follows:

\begin{itemize}

\item We formulate subclass-aware hybrid prototypes by blending dataset-specific visual prototypes with CONCH pathology text prototypes and matching them to multi-scale features from a frozen UNI encoder.

\item We propose an activation-balancing operator that performs class-wise power recalibration while exactly preserving each pixel’s original total foreground mass, thereby changing relative class allocation without creating or suppressing foreground evidence.

\item We propose a one-step graph diffusion mechanism that regularizes feature affinity with an attention-context discrepancy from late UNI blocks; this cue is treated as a soft structural proxy rather than an explicit boundary detector.

\item We distinguish direct stage-1 pseudo-mask evaluation from downstream stage-2 segmentation and evaluate the framework on BCSS-WSSS and LUAD-HistoSeg, with limitations of cross-backbone comparisons stated explicitly.

\end{itemize}
\begin{figure}[htbp]
    \centering
    \includegraphics[
        width=\columnwidth,
        trim=0 20 3 30,
        clip
    ]{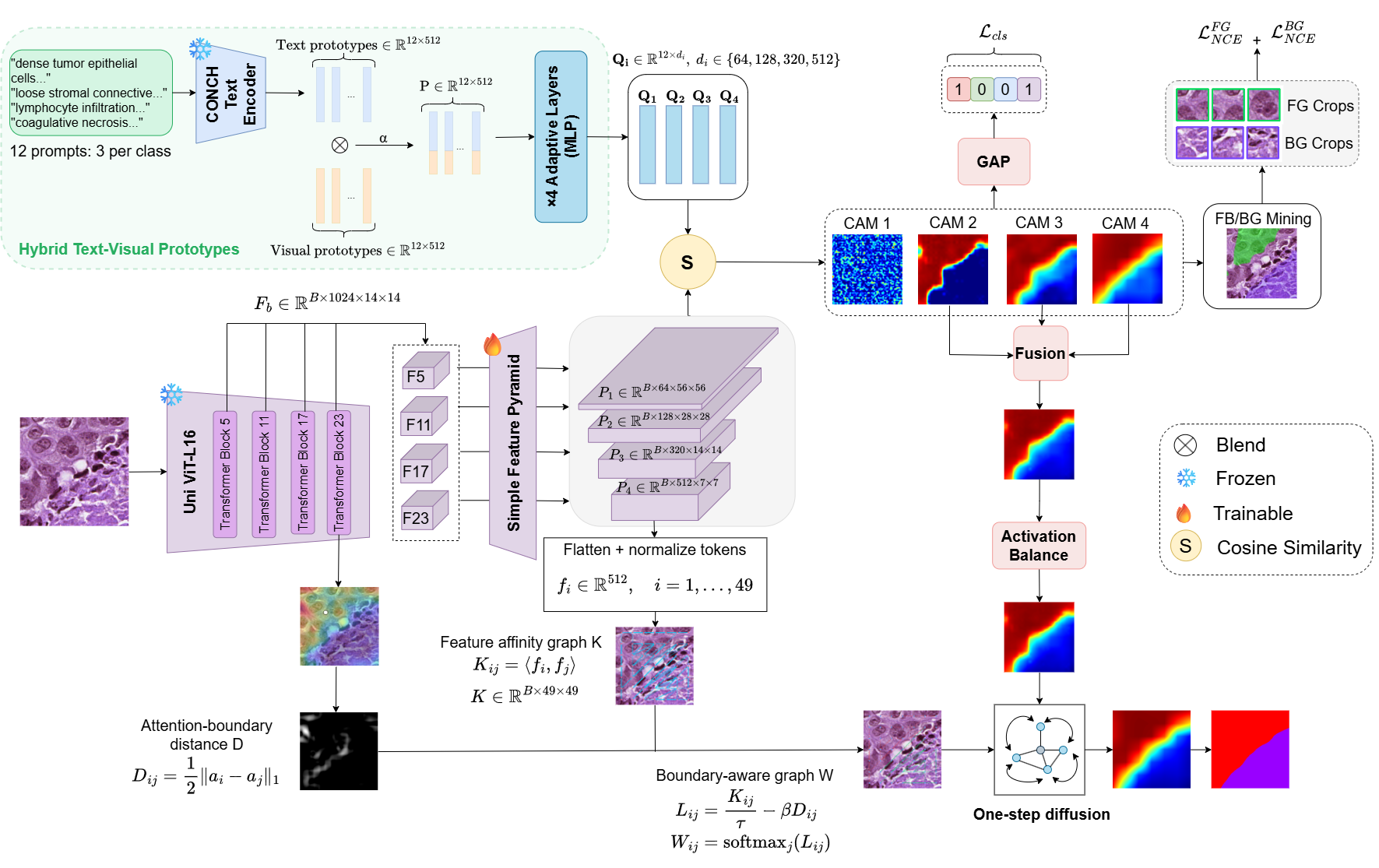}
    \caption{Overview of the ProBAG framework. Hybrid text--visual prototypes are matched with multi-scale features from a frozen pathology encoder to generate tissue CAMs, which are then fused, activation-balanced, and refined by boundary-aware graph diffusion to produce pseudo masks.}
    \label{fig:overview}
\end{figure}

\section{Methodology}
\textbf{Overview.} We study weakly supervised semantic segmentation (WSSS) in \\ histopathology, where each image $x$ has only an image-level multi-label annotation $y \in \{0, 1\}^C$ over parent tissue classes. Stage 1, which constitutes ProBAG, combines hybrid prototype CAM learning, activation-balanced CAM fusion, and Boundary-Aware Graph Diffusion (BAGD) to generate dense pseudo-masks without pixel-level supervision. Tables 2--4 evaluate these stage-1 masks directly. For the complete two-stage comparison in Table 1, the generated masks supervise a downstream Phikon-FPN model with LoRA; this stage-2 segmenter is an evaluation component rather than part of the proposed pseudo-mask generator. Stage-1 mask generation uses frozen UNI features and CONCH semantic guidance and does not rely on CRF or an external segmentation model.
\subsection{Hybrid Prototype CAM Learning}
\label{subsec:cam_learning}

A frozen UNI ViT-L/16 pathology foundation encoder $E$ extracts dense ViT tokens from the input image \cite{dosovitskiy2021imageworth16x16words,chen2024uni}. We take intermediate encoder blocks and transform them with a lightweight feature pyramid \cite{lin2017featurepyramidnetworksobject},
\begin{equation}
    \{F_{\ell}\}_{\ell=1}^{4}=\mathrm{FPN}(E(x)), \quad
    F_{\ell}\in\mathbb{R}^{B\times d_{\ell}\times H_{\ell}\times W_{\ell}},
\end{equation}
where lower levels retain finer spatial detail and the deepest level provides semantic features for graph refinement. Only the feature pyramid, prototype projections, and scale logits are trainable. The backbone roles are complementary: UNI supplies dense pathology features, CONCH supplies pathology-aligned text semantics, frozen MedCLIP is used only to encode CAM-mined regions for the contrastive auxiliary loss, and Phikon is reserved for the downstream stage-2 segmenter with LoRA. These models therefore occupy distinct functional roles, but the present study does not provide a controlled comparison against alternative pathology foundation models.

We represent each parent tissue class by several subclasses. Let $\mathcal{K}_c$
be the set of subclasses of parent class $c$, and let $\pi(k)$ map subclass $k$
to its parent. For each subclass, a visual prototype $v_k$ from the prototype bank is blended with a pathology text prototype $t_k$ obtained from CONCH ViT-B/16 \cite{lu2024avisionlanguage}:
\begin{equation}
    p_k =
    \frac{\alpha\bar t_k+(1-\alpha)\bar v_k}
    {\|\alpha\bar t_k+(1-\alpha)\bar v_k\|_2},
    \label{eq:hybrid_proto}
\end{equation}
where $\bar t_k$ and $\bar v_k$ are $\ell_2$-normalized and $\alpha$ controls
the fixed text--visual contribution. A level-specific projection head maps the
hybrid prototype to each pyramid dimension, yielding $p_{\ell,k}$.

At scale $\ell$, subclass CAM logits are computed by scaled cosine similarity
between normalized pixel features and projected prototypes:
\begin{equation}
    A_{\ell,k}(u)=s_{\ell}
    \left\langle
    \frac{F_{\ell}(u)}{\|F_{\ell}(u)\|_2},
    \frac{p_{\ell,k}}{\|p_{\ell,k}\|_2}
    \right\rangle .
    \label{eq:subclass_cam}
\end{equation}
Global average pooling gives image-level subclass logits $q_{\ell,k}$. Subclass
logits are merged into parent predictions,
\begin{equation}
    q_{\ell,c}^{\mathrm{par}}=\mathrm{Merge}(\{q_{\ell,k}:k\in\mathcal{K}_c\}),
\end{equation}

During stage-1 classification training, $\mathrm{Merge}$ is the arithmetic mean over subclasses, $q_{\ell,c}^{\mathrm{par}} = |\mathcal{K}_c|^{-1} \sum_{k \in \mathcal{K}_c} q_{\ell,k}$. During pseudo-mask generation, subclass CAMs are instead aggregated by a softmax-weighted sum over subclass responses (the implementation option named ``max'') before parent-level normalization.

and optimized only with image-level labels:
\begin{equation}
    \mathcal{L}_{\mathrm{cls}}
    =\sum_{\ell=1}^{4}\rho_{\ell}\,
    \mathrm{BCE}(q_{\ell}^{\mathrm{par}},y).
    \label{eq:cls_loss}
\end{equation}

\noindent\textbf{CAM-guided prototype regularization.} The deepest subclass CAMs are also used to mine foreground and background image
regions. A dynamic threshold selects foreground regions for active subclasses;
the complementary regions serve as background. These masked regions are encoded
by a frozen MedCLIP vision encoder \cite{wang2022medclipcontrastivelearningunpaired} and trained with
foreground/background InfoNCE losses \cite{oord2019representationlearningcontrastivepredictive} against the
corresponding tissue prototypes. The total objective is
\begin{equation}
    \mathcal{L}=\mathcal{L}_{\mathrm{cls}}+
    \lambda_{\mathrm{nce}}
    (\mathcal{L}_{\mathrm{FG}}+\mathcal{L}_{\mathrm{BG}}+
    \mu\mathcal{R}_{\mathrm{cam}}),
    \label{eq:total_loss}
\end{equation}
where $R_{\mathrm{cam}} = \mathrm{mean}(A_4)$, $\lambda_{\mathrm{nce}} = 0.05$, $\mu = 5 \times 10^{-4}$, and the InfoNCE temperature is $0.07$; the dynamic CAM threshold ratio is $0.5$.

\subsection{Activation-Balanced CAM Fusion}
\label{subsec:activation_balance}

For pseudo-mask generation, subclass CAMs are first merged into parent CAMs using the softmax-weighted subclass aggregation described above. Absent classes are masked using the image-level labels, and each parent CAM is independently normalized before resizing to the input resolution. We then fuse the intermediate and deep CAMs,
\begin{equation}
    S_c=\sum_{\ell\in\mathcal{M}}\omega_{\ell}A_{\ell,c}^{\mathrm{par}},
    \qquad \sum_{\ell\in\mathcal{M}}\omega_{\ell}=1,
    \label{eq:cam_fusion}
\end{equation}
where $\mathcal{M}$ denotes the selected CAM scales; in our implementation
$\mathcal{M}=\{2,3,4\}$, since the first-level CAM is spatially detailed but
less semantically stable. The fused CAMs are not directly comparable across
tissue classes, since some
tissues produce broader or stronger activations than others. We therefore
reshape class competition by class-wise power calibration while
preserving the original foreground mass $m(u)=\sum_j S_j(u)$:
\begin{equation}
    \widehat{S}_c(u)=m(u)
    \frac{(S_c(u)+\epsilon)^{\gamma_c}}
    {\sum_j(S_j(u)+\epsilon)^{\gamma_j}},
    \qquad
    \widetilde{S}_c=(1-\eta)S_c+\eta\widehat{S}_c .
    \label{eq:activation_balance}
\end{equation}
This recalibrates class competition while preserving the foreground activation
mass.

\subsection{Boundary-Aware Graph Diffusion}
\label{subsec:bagd}

The balanced CAMs are refined by BAGD,
which builds a graph from the deepest feature map $F_4$, following the broader
use of graph-based propagation in representation learning \cite{kipf2017semisupervisedclassificationgraphconvolutional}. Let
$f_i$ denote the normalized feature token at graph node $i$. Feature
affinity is computed as
\begin{equation}
    K_{ij}=\langle f_i,f_j\rangle,
    \qquad
    \mathcal{N}(i)=\{j:K_{ij}\ge\delta\}\cup\{i\}.
\end{equation}
Feature similarity alone may propagate scores across tissue interfaces. To reduce such leakage, we extract late-block self-attention profiles \cite{vaswani2023attentionneed} from the frozen encoder, pool them to the F4 graph resolution, and normalize them. We use the discrepancy between pooled profiles as an attention-context distance—a soft proxy for structural separation rather than an explicit boundary detector:
\begin{equation}
    D_{ij}=\frac{1}{2}\|a_i-a_j\|_1 .
    \label{eq:attn_distance}
\end{equation}
A large $D_{ij}$ indicates that two nodes participate in different global attention contexts. Subtracting $\beta D_{ij}$ lowers the diffusion probability across these contextually distinct edges, while $K_{ij}$ retains local appearance affinity. The boundary-aware diffusion graph is obtained by
\begin{equation}
    L_{ij}=K_{ij}/\tau-\beta D_{ij},\quad j\in\mathcal{N}(i),
    \qquad
    W_{ij}=\frac{\exp(L_{ij})}{\sum_{r\in\mathcal{N}(i)}\exp(L_{ir})}.
    \label{eq:graph_softmax}
\end{equation}
For each class, the balanced CAM is downsampled to graph resolution as
$s_c\in\mathbb{R}^{n}$ and refined by one residual diffusion step:
\begin{equation}
    z_c=(1-\lambda_c)s_c+\lambda_c W s_c .
    \label{eq:bagd}
\end{equation}
The interpolation weight $\lambda_c \in [0, 1]$ controls the residual smoothing strength. In the released BCSS configuration, $\lambda_c = (0.35, 0.35, 0.20, 0.45)$ for (TUM, STR, LYM, NEC), the graph resolution is $7 \times 7$, $\delta = 0.55$, $\tau = 0.10$, $\beta = 2$, and the attention profile is extracted from block 23. Refined maps are upsampled and converted to pseudo-masks by foreground argmax,
\begin{equation}
    \hat{Y}(u)=\arg\max_{c\in\{1,\ldots,C\}}Z_c(u).
    \label{eq:pseudo_mask}
\end{equation}
We deliberately use one residual diffusion step: the $(1-\lambda_c)s_c$ term retains the original CAM evidence while $\lambda_c W s_c$ applies a single context-constrained correction. Repeated propagation increasingly behaves as a low-pass filter and may erase small, irregular histological structures spanning only a few graph nodes.

\section{Experiments}
\subsection{Datasets}
We evaluate on two histopathology WSSS datasets: BCSS-WSSS \cite{10.1093/bioinformatics/btz083} and LUAD-HistoSeg \cite{han2022wsss4luadgrandchallengeweaklysupervised}. BCSS-WSSS contains 31,826 H\&E patches (train: 23,422, val: 3,418, test: 4,986) covering tumor (TUM), stroma (STR), lymphocyte (LYM), and necrosis (NEC). LUAD-HistoSeg has 17,291 patches (train: 16,678, val: 306, test: 307) covering tumor epithelium (TE), necrosis (NEC), lymphocytes (LYM), and tumor-associated stroma (TAS). Per standard WSSS, training uses only image-level labels; val/test sets use pixel masks.
\subsection{Implementation Details}
    \textbf{Stage-1 implementation.} Experiments were run on an RTX PRO 4000 (24 GB). Frozen UNI blocks 5, 11, 17, and 23 \cite{chen2024uni}  provide the four feature levels. Each parent class uses three visual/text subclasses, with fixed hybrid weight $\alpha = 0.6$. The released BCSS configuration uses AdamW with base learning rate $5 \times 10^{-5}$, weight decay $10^{-3}$, effective batch size 16, classification scale weights (0, 0.1, 1, 1), $\lambda_{\mathrm{nce}} = 0.05$, and validation-mIoU checkpoint selection. Pseudo-masks fuse scales 2--4 with weights (0.3, 0.3, 0.4) and use the activation-balance and BAGD settings reported above.

    \noindent\textbf{Stage-2 downstream evaluation.} For Table 1 only, stage-1 pseudo-masks supervise a Phikon ViT-B backbone with an FPN decoder. The Phikon backbone is frozen except for LoRA adapters (rank 8, $\alpha_{\mathrm{LoRA}} = 8$), and the model is trained for 80 epochs with batch size 16 and AdamW learning rate $6 \times 10^{-5}$ using noise-robust Bootstrapped CE (noise coefficient 0.3) plus Dice loss. The best checkpoint is selected by validation mIoU. Baselines use their official downstream implementations; therefore Table 1 compares complete systems rather than isolating pseudo-mask quality under a common stage-2 architecture.
\subsection{Comparison with State of the Art}
\setlength{\floatsep}{6pt plus 1pt minus 1pt}
\setlength{\textfloatsep}{6pt plus 1pt minus 1pt}
\setlength{\intextsep}{6pt plus 1pt minus 1pt}
\noindent\textbf{Quantitative Results.} Table 1 reports our retrained baseline runs using official implementations and the dataset splits/evaluation pipeline used in this study; the values are therefore not copied from the original papers and their ordering is protocol-dependent. This is particularly visible on LUAD-HistoSeg, where our retrained CAM baseline exceeds TPRO and PBIP. On BCSS-WSSS, ProBAG improves over PBIP by +4.13 mIoU, +1.52 mDice, and +4.40 FwIoU. On LUAD-HistoSeg, it improves over the strongest retrained baseline, CAM, by +2.15 mIoU, +1.38 mDice, and +1.84 FwIoU. These margins describe complete two-stage systems and may reflect both pseudo-mask quality and differences in foundation/downstream backbones; Tables 2–4 provide the more direct evidence for the proposed stage-1 components.

\begin{table}[ht]
\caption{Quantitative comparison of final segmentation performance on BCSS-WSSS and LUAD-HistoSeg. Results are from a single run for parity with baselines (ablations report mean$\pm$std). \textit{Note:} PBIP's LUAD results (marked with $^\dagger$) reflect stage-1 pseudo-mask evaluation, as we could not reproduce its stage-2.}
\label{tab:sota_comparison}
\scriptsize
\centering
\setlength{\tabcolsep}{3.0pt}
\renewcommand{\arraystretch}{0.95} 
\begin{tabular}{l|ccc|cccc}
\toprule
\multicolumn{8}{c}{\textbf{BCSS--WSSS}} \\
\midrule
Method & \multicolumn{3}{c|}{Metrics (\%)} & \multicolumn{4}{c}{Per-class IoU (\%)} \\
\cmidrule(lr){2-4} \cmidrule(lr){5-8}
 & mIoU & mDice & FwIoU & TUM & STR & LYM & NEC \\
\midrule
CAM~(CVPR'16)~\cite{zhou2015learningdeepfeaturesdiscriminative} 
& 63.32 & 77.29 & 68.53 & 72.98 & \underline{68.13} & 56.19 & 55.99 \\
Grad-CAM~(ICCV'17)~\cite{Selvaraju_2019} 
& 58.04 & 72.68 & 66.33 & 71.46 & 66.57 & 53.81 & 40.31 \\
Proto2Seg~(IPMI'23)~\cite{pan2023humanmachineinteractivetissueprototype} 
& 57.42 & 72.24 & -- & 63.25 & 58.28 & 53.27 & 54.89 \\
TPRO~(MICCAI'23)~\cite{Zhang2023TPROTW} 
& 65.50 & 78.90 & 69.42 & 77.40 & 65.10 & 56.20 & 63.40 \\
PBIP~(CVPR'25)~\cite{tang2025prototypebasedimagepromptingweakly} 
& \underline{69.03} & \underline{82.84} & \underline{71.93} & \underline{78.13} & 65.56 & \underline{62.11} & \textbf{70.31} \\
\textbf{Ours (ProBAG)} 
& \textbf{73.16} & \textbf{84.36} & \textbf{76.33} & \textbf{81.87} & \textbf{75.31} & \textbf{66.29} & \underline{69.14} \\
\midrule
\multicolumn{8}{c}{\textbf{LUAD--HistoSeg}} \\
\midrule
Method & \multicolumn{3}{c|}{Metrics (\%)} & \multicolumn{4}{c}{Per-class IoU (\%)} \\
\cmidrule(lr){2-4} \cmidrule(lr){5-8}
 & mIoU & mDice & FwIoU & TE & NEC & LYM & TAS \\
\midrule
CAM~(CVPR'16)~\cite{zhou2015learningdeepfeaturesdiscriminative} 
& \underline{74.26} & \underline{85.19} & \underline{73.32} & \underline{75.57} & \underline{78.08} & 73.69 & 69.69 \\
Grad-CAM~(ICCV'17)~\cite{Selvaraju_2019} 
& 71.19 & 83.13 & 71.89 & 75.92 & 68.89 & 72.09 & 67.86 \\
Proto2Seg~(IPMI'23)~\cite{pan2023humanmachineinteractivetissueprototype} 
& 61.64 & 75.67 & -- & 69.58 & 43.64 & 70.63 & 62.70 \\
TPRO~(MICCAI'23)~\cite{Zhang2023TPROTW} 
& 72.90 & 84.27 & 73.14 & 74.30 & 68.80 & \textbf{77.30} & \textbf{71.20} \\
PBIP~(CVPR'25)~\cite{tang2025prototypebasedimagepromptingweakly}$^\dagger$ 
& 66.05 & 79.39 & 62.55 & 60.70 & 71.80 & 72.50 & 59.20 \\
\textbf{Ours (ProBAG)} 
& \textbf{76.41} & \textbf{86.57} & \textbf{75.16} & \textbf{77.68} & \textbf{82.03} & \underline{74.94} & \underline{71.00} \\
\bottomrule
\end{tabular}
\end{table}

\begin{figure}[t]
    \centering
    \includegraphics[width=\columnwidth]{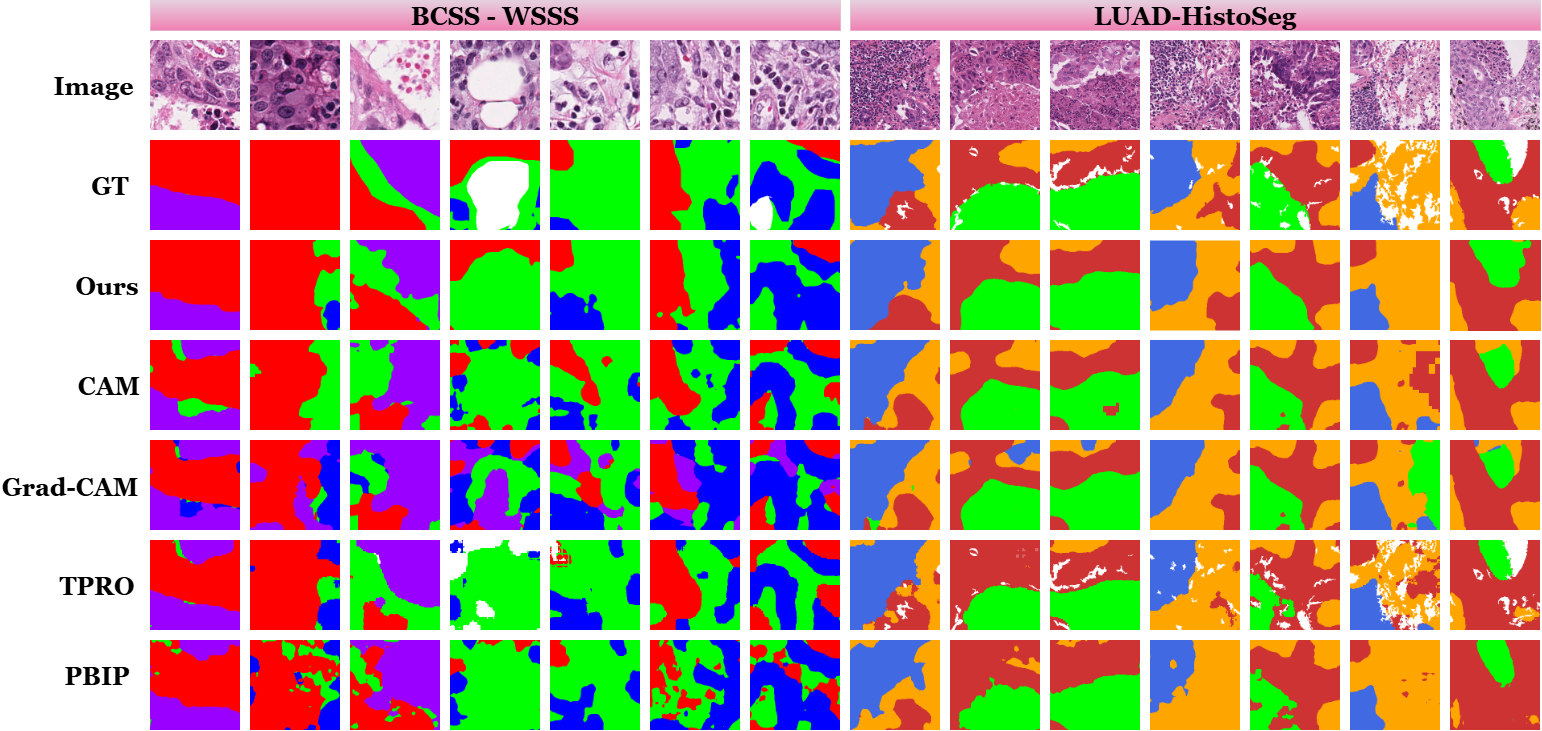}
    \caption{Qualitative comparison of final segmentation results on BCSS-WSSS and LUAD-HistoSeg test patches. GT denotes the ground truth segmentation.}
    \label{fig:qualitative_results}
\end{figure}

\noindent\textbf{Qualitative Results.} Figure~\ref{fig:qualitative_results} Figure 2 visually suggests that CAM, Grad-CAM, and PBIP can exhibit structural misalignment, diffuse interfaces, and pixel-level noise, whereas ProBAG often produces more coherent regions and sharper-looking transitions. Compared with TPRO, it also preserves several fine structures that are missing in the shown examples. These visualizations are consistent with the intended role of BAGD, but they are qualitative and do not substitute for a boundary-sensitive quantitative metric.
\subsection{Ablation Study}
\textbf{Effect of Main Components.}
Table \ref{tab:ablation_components} shows that pathology text semantics provide the dominant improvement: adding text prototypes raises mIoU from 68.72\% to 72.90\% ($+4.18$) and mDice from 81.22\% to 84.20\% ($+2.98$). Blending text and visual prototypes adds a smaller $+0.15$ mIoU and $+0.10$ mDice. Adding the final activation-balance and BAGD configuration raises performance by a further $+0.43$ mIoU and $+0.28$ mDice. Thus, the measured gains are complementary but not equal in magnitude; most of the improvement comes from pathology-aligned semantic prototypes, while calibration and graph refinement provide smaller corrections. The stage-1 pseudo-masks reach 73.48\% mIoU, close to and slightly above the 73.16\% stage-2 BCSS result. This difference may reflect pseudo-label noise and downstream optimization, but the current experiments do not establish a unique cause. Stage 1 is therefore reported as a CRF-free standalone output, while stage 2 is retained for comparison with two-stage systems.
\begin{table}[!htbp]
\centering
\caption{Effect of Main Components on initial pseudo masks for BCSS-WSSS.}
\label{tab:ablation_components}
\scriptsize
\setlength{\tabcolsep}{10pt}
\renewcommand{\arraystretch}{1.15}
\begin{tabular}{lcc}
\toprule
Setting & mIoU & mDice \\
\midrule
Visual baseline    & $68.72 \pm 0.67$ & $81.22 \pm 0.49$ \\
+ Text prototype   & $72.90 \pm 0.14$ & $84.20 \pm 0.10$ \\
+ Hybrid prototype & $73.05 \pm 0.09$ & $84.30 \pm 0.06$ \\
+ Hybrid + BAGD    & $\bm{73.48 \pm 0.20}$ & $\bm{84.58 \pm 0.14}$ \\
\bottomrule
\end{tabular}
\end{table}

\noindent\textbf{Effect of Text Encoder.} 
Within the same ProBAG stage-1 framework, Table~\ref{tab:ablation_text} shows that domain-specific language representations matter: replacing CLIP-style text features with CONCH increases mIoU by $+1.42\%$ and mDice by $+0.99\%$. This controlled comparison supports the use of pathology-aligned text semantics, while not constituting a broader comparison of visual foundation backbones.
\begin{table}[!htbp]
\centering
\caption{Effect of Text Encoders on initial pseudo masks for BCSS-WSSS.}
\label{tab:ablation_text}
\scriptsize
\setlength{\tabcolsep}{10pt}
\renewcommand{\arraystretch}{1.15}
\begin{tabular}{lcc}
\toprule
Text Encoder & mIoU & mDice \\
\midrule
No text      & $69.55 \pm 0.77$ & $81.82 \pm 0.54$ \\
CLIP         & $72.06 \pm 0.40$ & $83.59 \pm 0.27$ \\
CONCH        & $\bm{73.48 \pm 0.20}$ & $\bm{84.58 \pm 0.14}$ \\
\bottomrule
\end{tabular}
\end{table}

\noindent\textbf{Effect of Boundary Penalty.}
\begin{table}[!htbp]
\centering
\caption{Effect of the attention-boundary penalty $\beta$ on initial pseudo masks for BCSS-WSSS.}
\label{tab:ablation_beta}
\scriptsize
\setlength{\tabcolsep}{10pt}
\renewcommand{\arraystretch}{1.15}
\begin{tabular}{ccc}
\toprule
$\beta$ & mIoU & mDice \\
\midrule
0 & $73.39 \pm 0.22$ & $84.53 \pm 0.16$ \\
1 & $73.43 \pm 0.19$ & $84.55 \pm 0.14$ \\
2 & $73.48 \pm 0.20$ & $84.58 \pm 0.14$ \\
3 & $\bm{73.50 \pm 0.23}$ & $\bm{84.60 \pm 0.15}$ \\
\bottomrule
\end{tabular}
\end{table}
Setting $\beta = 0$ retains feature-only graph diffusion, whereas $\beta > 0$ adds the attention-context penalty. Across $\beta \in [0, 3]$, the region-based metrics vary by less than $0.2\%$ mIoU and $0.1\%$ mDice. In particular, the $\beta = 0$ to $\beta = 2$ gap is $0.09\%$ mIoU and $0.05\%$ mDice, smaller than the reported standard deviations. These results show that the overall diffusion pipeline is stable to $\beta$, but they do not by themselves establish a statistically separable benefit of the attention penalty on region metrics. We use $\beta = 2$ as a moderate fixed constraint rather than claiming it is uniquely optimal.

\subsection{Discussion and Limitations}
ProBAG combines established ingredients---prototype learning, multi-scale CAMs, and graph propagation---but targets two specific histopathology failure modes with explicit mechanisms: foreground-mass-preserving class recalibration and attention-context-regularized diffusion. The current evidence is strongest for pathology-aligned text prototypes; the additional region-metric gain of the attention penalty is modest and should be complemented by boundary-specific evaluation. Table~\ref{tab:sota_comparison} compares complete systems with different foundation and stage-2 backbones and is reported from single runs, so it cannot isolate pseudo-mask quality as cleanly as the stage-1 ablations. We also do not conduct an exhaustive comparison of pathology foundation models. These limitations motivate common-backbone evaluation, multi-seed final results, and dedicated boundary metrics.

\section{Conclusion}
We present ProBAG, a stage-1 weakly supervised histopathology segmentation framework that combines hybrid text–visual prototypes, foreground-mass-preserving activation balance, and attention-context-regularized graph diffusion. Results on BCSS-WSSS and LUAD-HistoSeg show consistent improvements over the retrained baselines, with ablations indicating that pathology-aligned text semantics account for the largest gain and graph refinement provides a smaller complementary correction. ProBAG produces pseudo-masks without CRF or an external segmentation model; a downstream Phikon-FPN is used only for the complete two-stage comparison. Future work will evaluate boundary-sensitive metrics, common-backbone controls, multi-seed final systems, stronger foundation-model boundary cues, and large-scale multi-center data.

%
%
%
\bibliographystyle{splncs04}
\bibliography{refs/mybib}

@misc{zhou2015learningdeepfeaturesdiscriminative,
      title={Learning Deep Features for Discriminative Localization}, 
      author={Bolei Zhou and Aditya Khosla and Agata Lapedriza and Aude Oliva and Antonio Torralba},
      year={2015},
      eprint={1512.04150},
      archivePrefix={arXiv},
      primaryClass={cs.CV},
      url={https://arxiv.org/abs/1512.04150}, 
}

@article{Selvaraju_2019,
   title={Grad-CAM: Visual Explanations from Deep Networks via Gradient-Based Localization},
   volume={128},
   ISSN={1573-1405},
   url={http://dx.doi.org/10.1007/s11263-019-01228-7},
   DOI={10.1007/s11263-019-01228-7},
   number={2},
   journal={International Journal of Computer Vision},
   publisher={Springer Science and Business Media LLC},
   author={Selvaraju, Ramprasaath R. and Cogswell, Michael and Das, Abhishek and Vedantam, Ramakrishna and Parikh, Devi and Batra, Dhruv},
   year={2019},
   month=Oct, pages={336–359}
}

@misc{kolesnikov2016seedexpandconstrainprinciples,
      title={Seed, Expand and Constrain: Three Principles for Weakly-Supervised Image Segmentation}, 
      author={Alexander Kolesnikov and Christoph H. Lampert},
      year={2016},
      eprint={1603.06098},
      archivePrefix={arXiv},
      primaryClass={cs.CV},
      url={https://arxiv.org/abs/1603.06098}, 
}

@misc{ahn2018learningpixellevelsemanticaffinity,
      title={Learning Pixel-level Semantic Affinity with Image-level Supervision for Weakly Supervised Semantic Segmentation}, 
      author={Jiwoon Ahn and Suha Kwak},
      year={2018},
      eprint={1803.10464},
      archivePrefix={arXiv},
      primaryClass={cs.CV},
      url={https://arxiv.org/abs/1803.10464}, 
}

@misc{ahn2019weaklysupervisedlearninginstance,
      title={Weakly Supervised Learning of Instance Segmentation with Inter-pixel Relations}, 
      author={Jiwoon Ahn and Sunghyun Cho and Suha Kwak},
      year={2019},
      eprint={1904.05044},
      archivePrefix={arXiv},
      primaryClass={cs.CV},
      url={https://arxiv.org/abs/1904.05044}, 
}

@misc{wang2020selfsupervisedequivariantattentionmechanism,
      title={Self-supervised Equivariant Attention Mechanism for Weakly Supervised Semantic Segmentation}, 
      author={Yude Wang and Jie Zhang and Meina Kan and Shiguang Shan and Xilin Chen},
      year={2020},
      eprint={2004.04581},
      archivePrefix={arXiv},
      primaryClass={cs.CV},
      url={https://arxiv.org/abs/2004.04581}, 
}

@misc{ru2022learningaffinityattentionendtoend,
      title={Learning Affinity from Attention: End-to-End Weakly-Supervised Semantic Segmentation with Transformers}, 
      author={Lixiang Ru and Yibing Zhan and Baosheng Yu and Bo Du},
      year={2022},
      eprint={2203.02664},
      archivePrefix={arXiv},
      primaryClass={cs.CV},
      url={https://arxiv.org/abs/2203.02664}, 
}

@misc{chen2022classreactivationmapsweaklysupervised,
      title={Class Re-Activation Maps for Weakly-Supervised Semantic Segmentation}, 
      author={Zhaozheng Chen and Tan Wang and Xiongwei Wu and Xian-Sheng Hua and Hanwang Zhang and Qianru Sun},
      year={2022},
      eprint={2203.00962},
      archivePrefix={arXiv},
      primaryClass={cs.CV},
      url={https://arxiv.org/abs/2203.00962}, 
}

@article{10.1093/bioinformatics/btz083,
    author = {Amgad, Mohamed and Elfandy, Habiba and Hussein, Hagar and Atteya, Lamees A and Elsebaie, Mai A T and Abo Elnasr, Lamia S and Sakr, Rokia A and Salem, Hazem S E and Ismail, Ahmed F and Saad, Anas M and Ahmed, Joumana and Elsebaie, Maha A T and Rahman, Mustafijur and Ruhban, Inas A and Elgazar, Nada M and Alagha, Yahya and Osman, Mohamed H and Alhusseiny, Ahmed M and Khalaf, Mariam M and Younes, Abo-Alela F and Abdulkarim, Ali and Younes, Duaa M and Gadallah, Ahmed M and Elkashash, Ahmad M and Fala, Salma Y and Zaki, Basma M and Beezley, Jonathan and Chittajallu, Deepak R and Manthey, David and Gutman, David A and Cooper, Lee A D},
    title = {Structured crowdsourcing enables convolutional segmentation of histology images},
    journal = {Bioinformatics},
    volume = {35},
    number = {18},
    pages = {3461-3467},
    year = {2019},
    month = {09},
    issn = {1367-4803},
    doi = {10.1093/bioinformatics/btz083},
    url = {https://doi.org/10.1093/bioinformatics/btz083},
    eprint = {https://academic.oup.com/bioinformatics/article-pdf/35/18/3461/48975491/bioinformatics_35_18_3461.pdf},
}

@article{Campanella2019ClinicalgradeCP,
  title={Clinical-grade computational pathology using weakly supervised deep learning on whole slide images},
  author={Gabriele Campanella and Matthew G. Hanna and Luke Geneslaw and Allen P. Miraflor and Vitor Werneck Krauss Silva and Klaus J. Busam and Edi Brogi and Victor E. Reuter and David S. Klimstra and Thomas J. Fuchs},
  journal={Nature medicine},
  year={2019},
  volume={25},
  pages={1301 - 1309},
  url={https://api.semanticscholar.org/CorpusID:196814162}
}

@misc{han2022wsss4luadgrandchallengeweaklysupervised,
      title={WSSS4LUAD: Grand Challenge on Weakly-supervised Tissue Semantic Segmentation for Lung Adenocarcinoma}, 
      author={Chu Han and Xipeng Pan and Lixu Yan and Huan Lin and Bingbing Li and Su Yao and Shanshan Lv and Zhenwei Shi and Jinhai Mai and Jiatai Lin and Bingchao Zhao and Zeyan Xu and Zhizhen Wang and Yumeng Wang and Yuan Zhang and Huihui Wang and Chao Zhu and Chunhui Lin and Lijian Mao and Min Wu and Luwen Duan and Jingsong Zhu and Dong Hu and Zijie Fang and Yang Chen and Yongbing Zhang and Yi Li and Yiwen Zou and Yiduo Yu and Xiaomeng Li and Haiming Li and Yanfen Cui and Guoqiang Han and Yan Xu and Jun Xu and Huihua Yang and Chunming Li and Zhenbing Liu and Cheng Lu and Xin Chen and Changhong Liang and Qingling Zhang and Zaiyi Liu},
      year={2022},
      eprint={2204.06455},
      archivePrefix={arXiv},
      primaryClass={eess.IV},
      url={https://arxiv.org/abs/2204.06455}, 
}

@inproceedings{Fang2023WeaklySupervisedSS,
  title={Weakly-Supervised Semantic Segmentation for Histopathology Images Based on Dataset Synthesis and Feature Consistency Constraint},
  author={Zijie Fang and Yang Chen and Yifeng Wang and Zhi Wang and Xiang Ji and Yongbing Zhang},
  booktitle={AAAI Conference on Artificial Intelligence},
  year={2023},
  url={https://api.semanticscholar.org/CorpusID:259735474}
}

@inproceedings{Zhang2023TPROTW,
  title={TPRO: Text-Prompting-Based Weakly Supervised Histopathology Tissue Segmentation},
  author={Shaoteng Zhang and Jianpeng Zhang and Yutong Xie and Yong Xia},
  booktitle={International Conference on Medical Image Computing and Computer-Assisted Intervention},
  year={2023},
  url={https://api.semanticscholar.org/CorpusID:263673303}
}

@misc{pan2023humanmachineinteractivetissueprototype,
      title={Human-machine Interactive Tissue Prototype Learning for Label-efficient Histopathology Image Segmentation}, 
      author={Wentao Pan and Jiangpeng Yan and Hanbo Chen and Jiawei Yang and Zhe Xu and Xiu Li and Jianhua Yao},
      year={2023},
      eprint={2211.14491},
      archivePrefix={arXiv},
      primaryClass={cs.CV},
      url={https://arxiv.org/abs/2211.14491}, 
}

@misc{tang2025prototypebasedimagepromptingweakly,
      title={Prototype-Based Image Prompting for Weakly Supervised Histopathological Image Segmentation}, 
      author={Qingchen Tang and Lei Fan and Maurice Pagnucco and Yang Song},
      year={2025},
      eprint={2503.12068},
      archivePrefix={arXiv},
      primaryClass={cs.CV},
      url={https://arxiv.org/abs/2503.12068}, 
}

@article{chen2024uni,
  title={Towards a General-Purpose Foundation Model for Computational Pathology},
  author={Chen, Richard J and Ding, Tong and Lu, Ming Y and Williamson, Drew FK and Jaume, Guillaume and Chen, Bowen and Zhang, Andrew and Shao, Daniel and Song, Andrew H and Shaban, Muhammad and others},
  journal={Nature Medicine},
  publisher={Nature Publishing Group},
  year={2024}
}

@article{lu2024avisionlanguage,
  title={A visual-language foundation model for computational pathology},
  author={Lu, Ming Y and Chen, Bowen and Williamson, Drew FK and Chen, Richard J and Liang, Ivy and Ding, Tong and Jaume, Guillaume and Odintsov, Igor and Le, Long Phi and Gerber, Georg and others},
  journal={Nature Medicine},
  pages={863–874},
  volume={30},
  year={2024},
  publisher={Nature Publishing Group}
}

@misc{wang2022medclipcontrastivelearningunpaired,
      title={MedCLIP: Contrastive Learning from Unpaired Medical Images and Text}, 
      author={Zifeng Wang and Zhenbang Wu and Dinesh Agarwal and Jimeng Sun},
      year={2022},
      eprint={2210.10163},
      archivePrefix={arXiv},
      primaryClass={cs.CV},
      url={https://arxiv.org/abs/2210.10163}, 
}

@misc{dosovitskiy2021imageworth16x16words,
      title={An Image is Worth 16x16 Words: Transformers for Image Recognition at Scale}, 
      author={Alexey Dosovitskiy and Lucas Beyer and Alexander Kolesnikov and Dirk Weissenborn and Xiaohua Zhai and Thomas Unterthiner and Mostafa Dehghani and Matthias Minderer and Georg Heigold and Sylvain Gelly and Jakob Uszkoreit and Neil Houlsby},
      year={2021},
      eprint={2010.11929},
      archivePrefix={arXiv},
      primaryClass={cs.CV},
      url={https://arxiv.org/abs/2010.11929}, 
}

@misc{lin2017featurepyramidnetworksobject,
      title={Feature Pyramid Networks for Object Detection}, 
      author={Tsung-Yi Lin and Piotr Dollár and Ross Girshick and Kaiming He and Bharath Hariharan and Serge Belongie},
      year={2017},
      eprint={1612.03144},
      archivePrefix={arXiv},
      primaryClass={cs.CV},
      url={https://arxiv.org/abs/1612.03144}, 
}

@misc{oord2019representationlearningcontrastivepredictive,
      title={Representation Learning with Contrastive Predictive Coding}, 
      author={Aaron van den Oord and Yazhe Li and Oriol Vinyals},
      year={2019},
      eprint={1807.03748},
      archivePrefix={arXiv},
      primaryClass={cs.LG},
      url={https://arxiv.org/abs/1807.03748}, 
}

@misc{vaswani2023attentionneed,
      title={Attention Is All You Need}, 
      author={Ashish Vaswani and Noam Shazeer and Niki Parmar and Jakob Uszkoreit and Llion Jones and Aidan N. Gomez and Lukasz Kaiser and Illia Polosukhin},
      year={2023},
      eprint={1706.03762},
      archivePrefix={arXiv},
      primaryClass={cs.CL},
      url={https://arxiv.org/abs/1706.03762}, 
}

@misc{kipf2017semisupervisedclassificationgraphconvolutional,
      title={Semi-Supervised Classification with Graph Convolutional Networks}, 
      author={Thomas N. Kipf and Max Welling},
      year={2017},
      eprint={1609.02907},
      archivePrefix={arXiv},
      primaryClass={cs.LG},
      url={https://arxiv.org/abs/1609.02907}, 
}

@misc{minaee2020imagesegmentationusingdeep,
      title={Image Segmentation Using Deep Learning: A Survey}, 
      author={Shervin Minaee and Yuri Boykov and Fatih Porikli and Antonio Plaza and Nasser Kehtarnavaz and Demetri Terzopoulos},
      year={2020},
      eprint={2001.05566},
      archivePrefix={arXiv},
      primaryClass={cs.CV},
      url={https://arxiv.org/abs/2001.05566}, 
}

@misc{lee2022weaklysupervisedsemanticsegmentation,
      title={Weakly Supervised Semantic Segmentation using Out-of-Distribution Data}, 
      author={Jungbeom Lee and Seong Joon Oh and Sangdoo Yun and Junsuk Choe and Eunji Kim and Sungroh Yoon},
      year={2022},
      eprint={2203.03860},
      archivePrefix={arXiv},
      primaryClass={cs.CV},
      url={https://arxiv.org/abs/2203.03860}, 
}

@misc{xie2022crosslanguageimagematching,
      title={Cross Language Image Matching for Weakly Supervised Semantic Segmentation}, 
      author={Jinheng Xie and Xianxu Hou and Kai Ye and Linlin Shen},
      year={2022},
      eprint={2203.02668},
      archivePrefix={arXiv},
      primaryClass={cs.CV},
      url={https://arxiv.org/abs/2203.02668}, 
}

@InProceedings{Van_Thai_2026_CVPR,
    author    = {Van Thai, Le and Nguyen, Tien Dat and Pham, Hoai Nhan and Thi, Lan Anh Dinh and Nguyen, Duy-Dong and Bui, Ngoc Lam Quang},
    title     = {UniSemAlign: Text-Prototype Alignment with a Foundation Encoder for Semi-Supervised Histopathology Segmentation},
    booktitle = {Proceedings of the IEEE/CVF Conference on Computer Vision and Pattern Recognition (CVPR) Workshops},
    month     = {June},
    year      = {2026},
    pages     = {6803--6813}
}

@misc{fu2025multimodalprototypealignmentsemisupervised,
      title={Multimodal Prototype Alignment for Semi-supervised Pathology Image Segmentation}, 
      author={Mingxi Fu and Fanglei Fu and Xitong Ling and Huaitian Yuan and Tian Guan and Yonghong He and Lianghui Zhu},
      year={2025},
      eprint={2508.19574},
      archivePrefix={arXiv},
      primaryClass={cs.CV},
      url={https://arxiv.org/abs/2508.19574}, 
}

\end{document}